\documentclass[letterpaper, 10 pt, conference]{ieeeconf}
\IEEEoverridecommandlockouts
\usepackage[usenames,dvipsnames]{xcolor}
\usepackage{graphics} %
\usepackage{graphicx, epstopdf}
\usepackage{amsmath,mathtools} %

\usepackage{amssymb}  %
\usepackage[nohyperlinks, nolist]{acronym}
\usepackage[pdftex, pdfstartview={FitV}, pdfpagelayout={TwoColumnLeft},bookmarksopen=true,plainpages = false, colorlinks=true, linkcolor=black, citecolor = black, urlcolor = black,filecolor=black , pagebackref=false,hypertexnames=false, plainpages=false, pdfpagelabels ]{hyperref}
\usepackage{balance}
\usepackage[capitalize]{cleveref}
\usepackage[sort,compress]{cite}
\usepackage[normalem]{ulem}
\usepackage[utf8]{inputenc}
\usepackage[T1]{fontenc}
\usepackage{mathtools}
\usepackage{xcolor}
\usepackage[thinc]{esdiff}
\usepackage[font=scriptsize]{caption}
\usepackage{subcaption}
\usepackage{paralist}
\usepackage{soul}

\title{\LARGE \bf
Airflow-Inertial Odometry for Resilient State Estimation
on Multirotors}

\author{Andrea Tagliabue and Jonathan P.\ How%
\thanks{A.~Tagliabue and J.~P.~How are with the Department of Aeronautics and Astronautics, Massachusetts Institute of Technology. \tt\{atagliab, jhow\}@mit.edu}%
}

\begin{document}

\maketitle
\thispagestyle{empty}
\pagestyle{empty}

\newacro{MAV}{Micro Aerial Vehicle}
\newacro{UKF}{Unscented Kalman filter}
\newacro{EKF}{extended Kalman filter}
\newacro{PF}{particle filter}
\newacro{LSTM}{Long Short-Term Memory}
\newacro{RNN}{recurrent neural network}
\newacro{CNN}{convolutional neural network}
\newacro{CoM}{center of mass}
\newacro{USQUE}{Unscented Quaternion estimator}
\newacro{UT}{Unscented Transformation}
\newacro{MSE}{Mean Squared Error}
\newacro{GP}{Gaussian Process}
\acrodefplural{GP}[GPs]{Gaussian Processes}
\newacro{MAP}{Maximum a Posterior}
\newacro{IMU}{Inertial Measurement Unit}
\newacro{RMSE}{Root-Mean-Square error}
\newacro{RTE}{Relative Translation error}
\newacro{AIO}{Airflow-Inertial Odometry}
\newacro{VO}{Visual-Odometry}
\newacro{RBF}{radial basis function}

\begin{abstract}
We present a dead reckoning strategy for increased resilience to position estimation failures on multirotors, using only data from a low-cost IMU and novel, bio-inspired airflow sensors. The goal is challenging, since low-cost IMUs are subject to large noise and drift, while 3D airflow sensing is made difficult by the interference caused by the propellers and by the wind. Our approach relies on a deep-learning strategy to interpret the measurements of the bio-inspired sensors, a map of the wind speed to compensate for position-dependent wind, and a filter to fuse the information and generate a pose and velocity estimate. Our results show that the approach reduces the drift with respect to IMU-only dead reckoning by up to an order of magnitude over 30 seconds after a position sensor failure in non-windy environments, and it can compensate for the challenging effects of turbulent, and spatially varying wind. 
\end{abstract}

\section{INTRODUCTION}
\label{sec:intro}
Resilient odometry estimation for robots navigating in GPS- and perceptually-degraded environments remains a challenging task. Motion blur, featureless-conditions or jamming can cause temporary failures in vision, lidar, or GPS-based position estimation systems, with catastrophic consequences for inherently unstable platforms such as multirotors. For these reasons, the potentially unreliable measurements from GPS, lidars or cameras are usually fused with the data from an \ac{IMU} \cite{ye2019tightly, bloesch2015robust, hausman2016self}, whose acceleration and angular velocity measurements can be used to perform dead reckoning\footnote{Dead reckoning: the task of estimating pose/velocity using only velocity or acceleration measurements, in our context because a failure of the source of position measurements (e.g. GPS, Visual-Odometry) occurs.} between the temporary failures of the position sensors/estimators. Measurements from consumer-grade \acp{IMU}, however, are typically corrupted by drifting biases, calibration errors and noise \cite{jang2020analysis}, causing large position drifts when integrated, and making \ac{IMU}-only dead reckoning reliable for a short amount of time. 

In this work we propose to improve \ac{IMU}-based dead reckoning via 3D airflow data, as a lightweight and cost-effective way to increase the resilience of multirotors to position estimation failures.
This idea is inspired by nature, and the way that bats \cite{sterbing2011bat} and \textit{Drosophila} \cite{fuller2014flying} use hair-like airflow receptors to obtain velocity feedback during flight. We build upon our previous work, where we combined measurements obtained from novel, bio-inspired sensors \cite{kim2020whisker} with a deep-learning strategy \cite{tagliabue2020touch} to estimate the 3D velocity of the robot. Different from our previous work \cite{tagliabue2020touch}, where the focus is on the estimation of drag and other forces assuming position information is always available, %
we now fuse airflow measurements with IMU data 
to compensate for time-varying biases and other sources of estimation errors, reducing position and velocity drifts experienced in IMU-only dead reckoning.  While the idea of combining IMU and 3D airflow data for dead reckoning has been explored in the literature of fixed-wing aircraft (e.g., \cite{berman1996role, leutenegger2012low, gebre2004design, johansen2015estimation}) or blimps (e.g., \cite{muller2012probabilistic, muller2013efficient}), it has not found application for multirotors, in part because of the large levels of noise \cite{prudden2018measuring} caused by the close proximity of the airflow sensors to the propellers. %
\begin{figure}[t]
\centering
\begin{subfigure}{.4\columnwidth}
  \centering %
  \includegraphics[width=\textwidth, trim=30 200 130 100, clip]{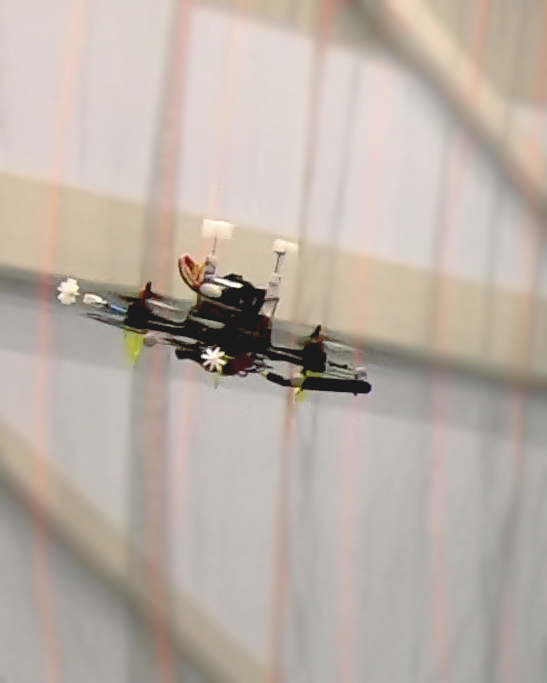}
\end{subfigure}%
\begin{subfigure}{.6\columnwidth}
  \centering
  \includegraphics[width=.9\textwidth, trim=30 25 30 45, clip]{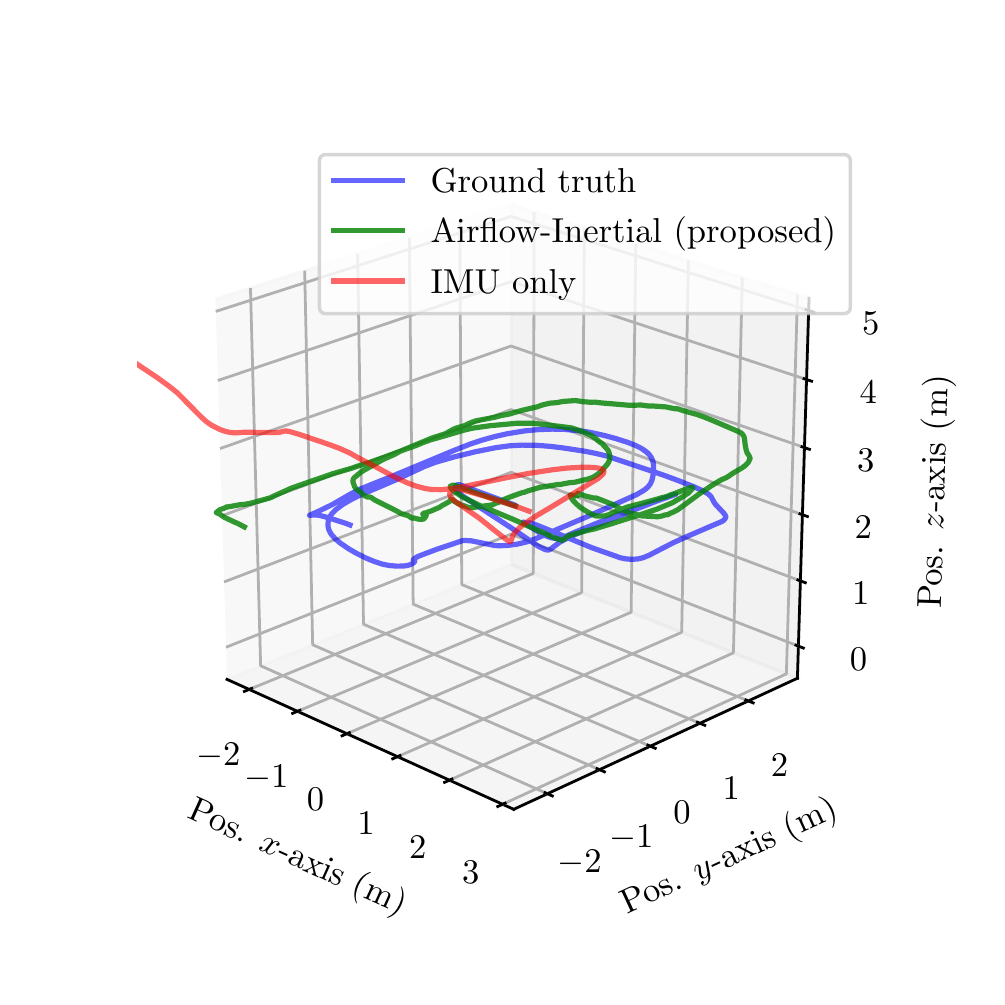}
\end{subfigure}
    \caption{Multicopter equipped with whisker-inspired airflow sensors (left) used to perform a 30s long dead reckoning (right), without relying on any other sensing source beyond a low-cost IMU. We compare our strategy with a 13s long trajectory obtained via IMU-only dead reckoning (in red), which visibly drifts away. }
    \label{fig:intro_figure} \vspace*{-0.25in}
\end{figure}

In order to deploy our strategy in windy conditions, in which the measured relative airflow differs from the velocity of the robot, 
we estimate and compensate for the effects of wind. 
While wind speed is typically assumed to be constant (independent of position and time, e.g., \cite{leutenegger2012low}), we instead assume the presence of a stationary wind field, meaning that the wind can vary spatially, but does not change with time. This assumption enables our approach to better handle complex environments such as an urban canopy \cite{ware2016analysis}. The stationary wind field is accounted for by generating a probabilistic representation of the mean wind field and then integrating this representation into our estimation scheme. 

Results from data collected experimentally show that the proposed \ac{AIO} strategy achieves one order of magnitude reduction in position errors (RMSE, drift) compared to IMU-only dead reckoning, during a 30-seconds pose and velocity measurements failure and in an environment with no wind. We additionally show that our strategy can compensate for the effects of turbulent, stationary wind produced by an array of leaf-blowers, reducing the drift with respect to IMU-only dead reckoning.

\textbf{Contributions}: present and evaluate an approach:
\begin{inparaenum}[i)]
    \item for IMU and airflow-based dead reckoning using novel, bio-inspired 3D airflow sensors;
    \item to relax the common assumption of constant wind in the environment, and capable of compensating the effects of challenging, turbulent wind.
\end{inparaenum}

\section{RELATED WORK}
\label{sec:related_works}
\begin{inparaenum}[]

    \item \textbf{Sensor fusion strategies for increased resilience of multirotors}: A common approach adopted to guarantee resilience to state estimation failures on multirotors is to combine data from multiple, different exteroceptive sensors with complementary failure modalities while employing a kinematic model of the robot \cite{lynen2013robust, hausman2016self, 
    li2019imu, santamaria2020towards}. Most of these approaches, however, rely on IMU data as main proprioceptive sensor, and may be subject to drift in case of simultaneous failure of their exteroceptive inputs.
    The works in \cite{svacha2020imu, svacha2019imu, svacha2019inertial, leishman2014quadrotors, martin2010true, kai2017nonlinear}, instead, fuse IMU data with thrust measurements via an accurate drag model of a multirotor, showing promising estimation performance without relying on exteroceptive signals. Contrary to our work, however, these strategies rely on a drag and dynamics model, which can be hard to obtain (e.g., may require specific trajectories to observe all required parameters) and can vary over time (e.g., during a package delivery) and vehicle/wind orientation.
    Similar to our work, \cite{zahran2018new} presents a strategy for airflow-inertial dead reckoning on a multirotor, but it does not account for wind effects.

    \item \textbf{IMU-only dead reckoning with pseudo measurements}: Recent works have shown promising result for IMU-only  dead reckoning, combining inertial data with virtual/pseudo measurements. %
    The works \cite{liu2020tlio, chen2018ionet, yan2019ronin}, for example, assume that the sensor is mounted on a pedestrian, and use deep-learning to generate relative displacement measurements by recognizing typical walking phases.
    The works \cite{lew2019contact, wuusing} generate instead zero-velocity pseudo measurement updates by detecting contacts of a UAV with the environment. These approaches are not applicable in our context since they make specific assumptions about the motion of the robot/sensor.

    \item \textbf{Relative airflow sensing}: The velocity of the airflow surrounding a multirotor can be estimated by detecting its inertial effects, such as the accelerations caused by the drag force,
    via Bayesian filtering
    \cite{demitrit2017model, sikkel2016novel, rodriguez2016wind} or learning-based approaches \cite{shi2019neural, allison2019estimating, marton2019hybrid, tagliabue2020touch}. These techniques require position or velocity information, as the airflow would be otherwise unobservable or hard to distinguish from the effects of other forces.
    Strategies to directly sense airflow onboard utilize a variety of sensors, such as ultrasonic \cite{hollenbeck2018wind}, whisker-like \cite{deer2019lightweight} or pressure-based \cite{bruschi2016wind}, and deep-learning techniques are employed in their calibration process \cite{calia2008multi, lerro2012development, donofrio2020all}. Contrary to our work, most of these approaches focus on estimating the magnitude of airflow velocity, but not its 3D direction.

    \item \textbf{Wind mapping via wind field estimation}: Learning-based strategies are a popular tool for wind mapping and estimation of the wind field. Among the non-parametric methods, \acp{GP} have found applications to estimate the mean wind field \cite{lawrance2010simultaneous, yang2017real} and to make forecasts \cite{chen2013short}.
    Regarding supervised methods, \cite{achermann2019learning} uses a \ac{CNN} to predict the wind field given the 3D geometry of the environment; \cite{erichson2020shallow} proposes a shallow neural network to estimate time-varying flows. %
    Approaches that more explicitly attempt to model the spatial distribution of the wind are presented by \cite{rodriguez2016wind}, where the relationship between wind speed and altitude is assumed to follow a Weibull distribution, while  \cite{langelaan2012wind} employs a polynomial model whose parameters can be estimated online. 
    
\end{inparaenum}

\section{APPROACH OVERVIEW}
\label{sec:overview}
\begin{figure}
    \centering
    \vspace*{0.05in}
    \includegraphics[width=\columnwidth]{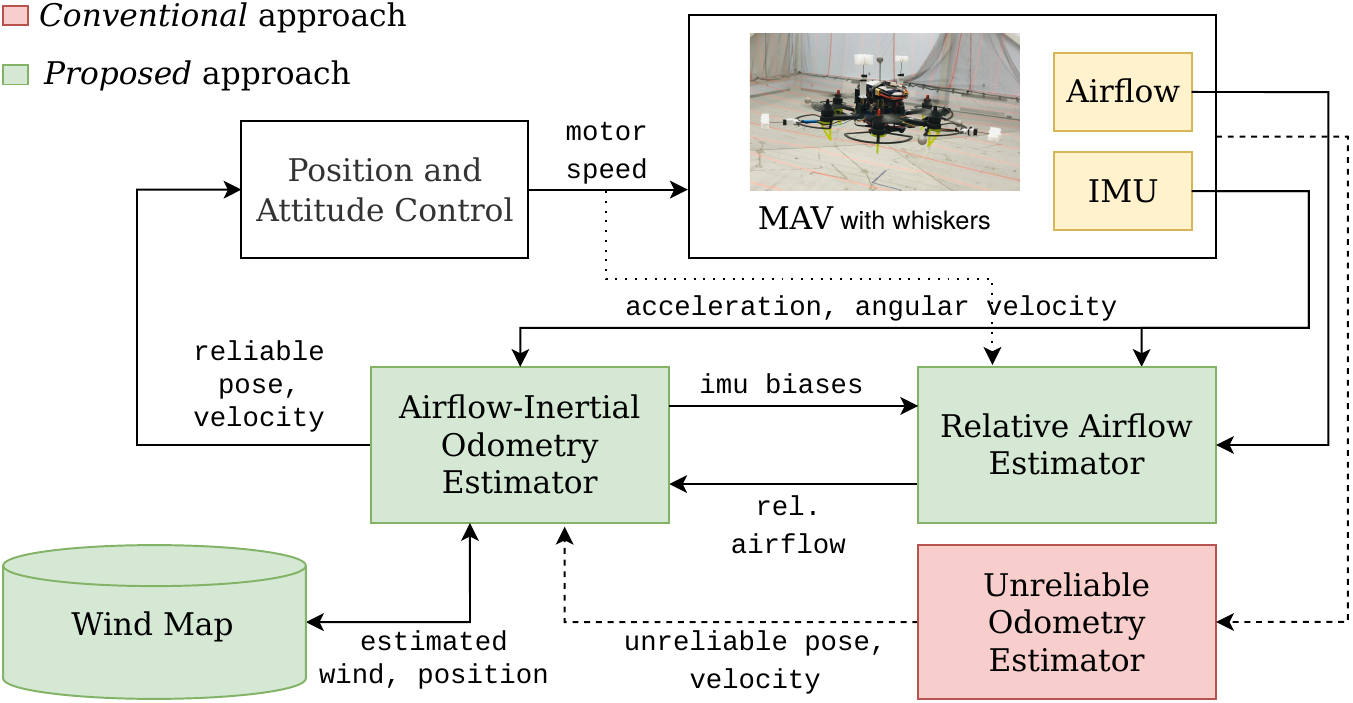}
    \vspace*{-0.2in}
    \caption{System diagram of the proposed approach. The \acf{AIO} estimator performs dead reckoning via airflow and inertial data whenever a failure of the Unreliable Odometry Estimator occurs. It can also compensate for the effects of position-dependent wind (static wind field) via a wind map.}
    \label{fig:system_diagram}
    \vskip-2ex
\end{figure}

The proposed approach relies on multiple components, shown in the system diagram in \cref{fig:system_diagram}.
\begin{inparaenum}[(a)]
    \item \textbf{Unreliable Odometry Estimator} provides pose information (and may additionally provide velocity measurements), and can be based on GPS, \ac{VO} or any other existing pose estimation system typically available onboard a multirotor. It is "unreliable" because it may suddenly fail by not outputting any estimate. 
    \item \textbf{Relative Airflow Estimator} provides a 3D estimate of the relative airflow surrounding the multirotor from the measurements of the bio-inspired airflow sensors (\cref{sec:wind_sensing}).
    \item \textbf{Airflow-Inertial Odometry Estimator}'s main goal is to provide an always-available pose estimate to the controllers via dead reckoning, by combining IMU data with airflow measurements and with the output of the Unreliable Odometry Estimator (whenever available). If the approach is deployed in environments with stationary wind (e.g. wind can change in position), then it additionally requires and integrates a pre-computed wind map (which, we note, is not needed for constant wind environments) (\cref{sec:odometry}).
    \item \textbf{Wind Map} provides an estimate of the velocity of the wind at a given position, to compensate for the presence of stationary wind fields (\cref{sec:wind_mapping}). %
\end{inparaenum}

\section{RELATIVE AIRFLOW ESTIMATION}
\label{sec:wind_sensing}
This section presents the strategy adopted to estimate the 3D relative airflow surrounding the robot using whisker-like sensors. This approach is based on our previous works \cite{kim2020whisker, tagliabue2020touch, paris2020control} and is summarized here for completeness. 

\begin{figure}
\centering
\begin{subfigure}{.6\columnwidth}
  \centering
  \includegraphics[width=0.9\columnwidth]{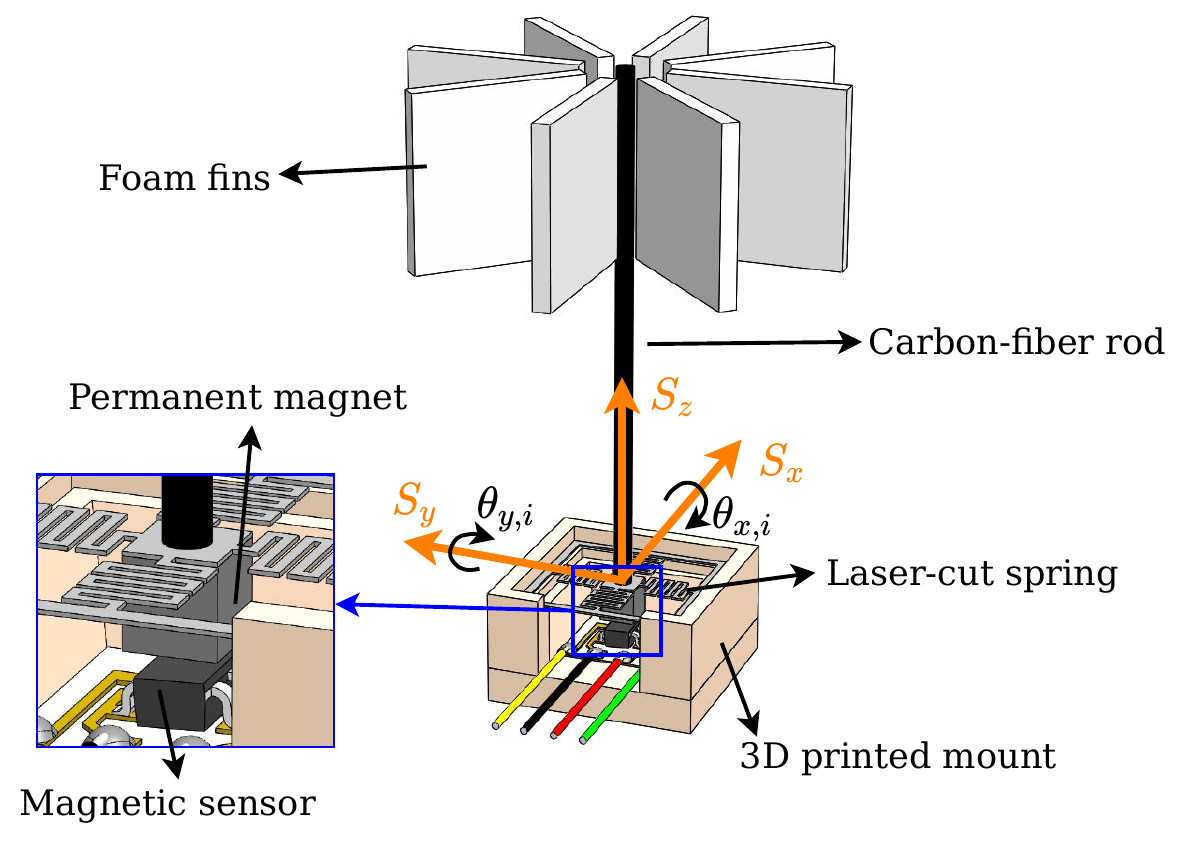}
\end{subfigure}%
\begin{subfigure}{.3\columnwidth}
  \centering
  \includegraphics[width=0.8\columnwidth, trim = 0 0 0 100, clip]{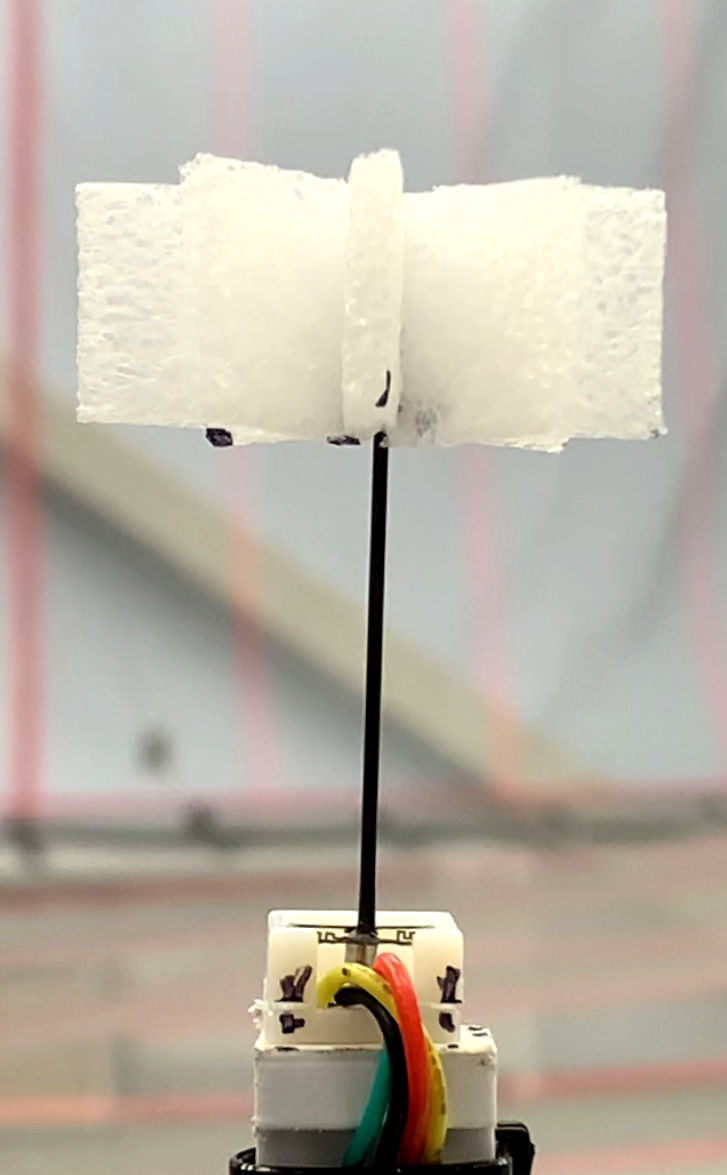}
\end{subfigure}
\vspace*{-0.1in}
    \caption{Illustration from our previous work \cite{tagliabue2020touch} and image of an actual whisker-like, bio-inspired sensor used to estimate the airflow surrounding the robot.}
    \label{fig:sensor_design}
    \vskip-2ex
\end{figure}

\paragraph{Sensor design}
In order to accurately measure the relative 3D airflow surrounding a multirotor we use four whisker-inspired sensors, shown in \cref{fig:intro_figure} and \cref{fig:sensor_design}. The chosen sensor design, based on \cite{kim2020whisker}, is inspired by nature, and the 3D flow-sensing capabilities found in whiskers from animals like rats \cite{yan2016mechanical} or seals \cite{dehnhardt1998seal}.
Our design uses foam fins mounted on a rod hinged on a torsional spring. When the $i$-th sensor is subjected to airflow, the drag force on the fins causes a rotation $\boldsymbol{\theta}_i = [\theta_{x,i}, \theta_{y,i}]^{T}$ of the rod. Such rotation is detected with a hall-effect sensor and a magnet mounted below the spring, which rotates with the rod. 

\paragraph{Relative airflow estimator}
The goal of the Relative Airflow Estimator, based on \cite{tagliabue2020touch}, is to produce a relative airflow estimate $\hat{\mathbf{v}}_\infty$, expressed in the body frame $B$ of the robot, from the measurements of the four sensors $\boldsymbol{\theta}= \{\boldsymbol{\theta}_1, \dots, \boldsymbol{\theta}_4\}$. 
The estimator is a \ac{LSTM} deep neural network, which can take into account hard-to-model effects such as the aerodynamic disturbances caused by the induced velocity from the propellers and the body of the robot \cite{prudden2018measuring, ventura2018high}, and time-dependent effects \cite{lipton2015critical, goodfellow2016deep} related to  the dynamics of the sensors and of the airflow. To provide the information needed to compensate for these effects, we augment the input of the network with the de-biased acceleration and angular velocity measurement from the IMU, and the normalized throttle commanded to the six propellers\footnote{We note that in many aerial platforms the commanded throttle slightly changes with the battery voltage. Including the battery voltage as an additional input of the LSTM could therefore further improve accuracy.}. The usefulness of the selected signal is confirmed by the ablation study presented in \cref{sec:evaluation}.

\section{AIRFLOW-INERTIAL ODOMETRY ESTIMATOR}
\label{sec:odometry}
This section presents the strategy utilized for airflow-inertial dead-reckoning, fusing IMU, airflow and odometry data from the Unreliable Odometry Estimator (when available). The proposed estimator can additionally compensate for the effects of stationary wind by utilizing a pre-generated wind map. Optionally, if a wind map does not exist yet, this estimator can be used to collect the wind-estimates required to generate (offline) the wind map. This step requires position information (from the Unreliable Odometry Estimator) to be available, as wind is otherwise unobservable \cite{leutenegger2012low}. 

\textbf{Notation:} a vector $\prescript{}{B}{\mathbf{r}}$ described in the body-fixed reference frame $B$ can be expressed in the inertial frame $W$ via $\prescript{}{W}{\mathbf{r}} = {\mathbf{R}_{W}^{B}}\prescript{}{B}{\mathbf{r}}$. Following \cite{barfoot2017state}, $\mathbf{r}^\wedge$ denotes the $3 \times 3$ skew-symmetric matrix associated with $\mathbf{r}$, and $(\mathbf{r}^\wedge)^\vee = \mathbf{r}$ the inverse operation. The subscript $\mathbf{r}_m$ denotes a measurement.

\subsection{Prediction model and state}
\paragraph{IMU model}
The prediction step is purely driven by the acceleration $\prescript{}{B}{\mathbf{a}}_m$ and the angular velocity $\prescript{}{B}{\boldsymbol{\omega}}_m$ measured by the IMU, where $B$ denotes the body reference frame, fixed to the IMU. These measurements are corrupted by white noise $\boldsymbol{\eta}_i  \sim \mathcal{N}(\boldsymbol{0},\,\boldsymbol{\Sigma}_i)$, and by slowly drifting biases $\mathbf{b}_i$ (with $i=a$ for the accelerometer and $i=g$ for the gyro): 
\begin{equation}
    \prescript{}{B}{\boldsymbol{\omega}} = \prescript{}{B}{\boldsymbol{\omega}}_m - \mathbf{b}_g + \boldsymbol{\eta}_{g}, \hspace{6pt}
    \prescript{}{B}{\mathbf{a}} = \prescript{}{B}{\mathbf{a}}_m - \mathbf{b}_a + \boldsymbol{\eta}_a \\ 
\end{equation}
The vectors $\prescript{}{B}{\boldsymbol{\omega}}$ and  $\prescript{}{B}{\mathbf{a}}$ denote the true angular velocity and the true acceleration of the system, respectively. The biases evolve according to a random walk process %
\begin{equation}
\begin{split}
    \dot{\mathbf{b}}_a = \boldsymbol{\eta}_{ba}, \hspace{4pt} \dot{\mathbf{b}}_g = \boldsymbol{\eta}_{bg} \\
    \end{split}
    \label{eq:bias_rw}
\end{equation}
where $\boldsymbol{\eta}_{ba}$, $\boldsymbol{\eta}_{bg}$ are zero-mean, Gaussian random variables. 
\paragraph{Kinematic model}
The position $\prescript{}{W}{\mathbf{p}}$, velocity $\prescript{}{W}{\mathbf{v}}$ and attitude of the robot ${\mathbf{R}_{W}^{B}}$ expressed in an inertial reference frame $W$ can be obtained from IMU measurements
\begin{equation}
\begin{split}
    \prescript{}{W}{\dot{\mathbf{p}}} &= \prescript{}{W}{\mathbf{v}} \\
    \prescript{}{W}{\dot{\mathbf{v}}} &= \prescript{}{W}{\mathbf{g}} + {\mathbf{R}_{W}^{B}}(\prescript{}{B}{\mathbf{a}}_m - \mathbf{b}_a + \boldsymbol{\eta}_a) \\
    \dot{\mathbf{R}}_{W}^{B} &= \mathbf{R}_{W}^{B} (\prescript{}{B}{\boldsymbol{\omega}}_m - \mathbf{b}_g + \boldsymbol{\eta}_{g})^\wedge
    \label{eq:kinematics:cont}
\end{split}
\end{equation}
and $\prescript{}{W}{\mathbf{g}}$ denotes the gravity vector. 
\paragraph{Wind model}
The three-dimensional true wind $\prescript{}{W}{\mathbf{w}}$ experienced by the robot is described via 
\begin{equation}
    \prescript{}{W}{{\mathbf{w}}} = \hat{\mathcal{M}}(\prescript{}{W}{\mathbf{p}}) + \boldsymbol{\eta}_m +  \mathbf{\mathbf{e}}_w 
    \label{eq:wind}
\end{equation}
where $\hat{\mathcal{M}}(\cdot)$ represents the estimated map of the wind, which outputs the mean field as a function of estimated position. The wind map is represented via a \acf{GP}, and is a continuous, differentiable function, subject to position-dependent uncertainty  $\boldsymbol{\eta}_m\sim\mathcal{N}(\boldsymbol{0},\,\boldsymbol{\Sigma}_\mathcal{M}(\mathbf{p}))$. The term $\mathbf{\mathbf{e}}_w$ denotes errors in the wind estimated by the map, due to empty map or drifts in the estimated position. We assume that changes in $\mathbf{\mathbf{e}}_w$ are driven by a random process
\begin{equation}
    \dot{\mathbf{e}}_w = \boldsymbol{\eta}_w, \hspace{4pt} \boldsymbol{\eta}_w \sim \mathcal{N}(\boldsymbol{0}, \boldsymbol{\Sigma}_w).
    \label{eq:wind_dynamics}
\end{equation}
If the wind map is empty and position information is available, $\mathbf{\mathbf{e}}_w$ can be used to estimate the wind by setting a positive covariance $\boldsymbol{\Sigma}_w$. If the wind map is available, we assume instead $\boldsymbol{\Sigma}_w = \boldsymbol{0}$, which reflects the assumption that errors due to position drifts do not change over time.

\paragraph{Continuous time prediction model and state}
The continuous time prediction model is obtained by combining the kinematic equations \cref{eq:kinematics:cont}, with the dynamics of the biases defined in \cref{eq:bias_rw} and the wind map error dynamics defined in equation \cref{eq:wind_dynamics}. 
The full state is then given by:  
\begin{equation}
    \mathbf{x}= \{\prescript{}{W}{\mathbf{p}}, \prescript{}{W}{\mathbf{v}}, \boldsymbol{\phi}, \prescript{}{B}{{\mathbf{b}_a}}, \prescript{}{B}{{\mathbf{b}_g}},   \prescript{}{W}{\mathbf{e}_w}\}
\end{equation}
where we express the attitude using the error-state representation $\boldsymbol{\phi}=\log({\mathbf{R}_{W}^{B}}_\text{ref}{\mathbf{R}_{W}^{B}}^{-1})^\vee$, in order to reduce effects caused by linearization errors, where $\boldsymbol{\phi} \in \mathbb{R}^3$ and the associated rotation matrix is obtained as $\exp(\boldsymbol{\phi}^\wedge) \in SO(3)$.  
The continuous-time process noise vector is given by the vectors 
$
    \boldsymbol{\eta} = \{\boldsymbol{\eta}_a, \boldsymbol{\eta}_g, \boldsymbol{\eta}_{ba}, \boldsymbol{\eta}_{bg}, \boldsymbol{\eta}_w\}
$.

\subsection{Measurement updates}
\paragraph{Airflow sensors measurement model} 
The Relative Airflow Estimator is considered as a new sensor, which produces a relative airflow measurements ${\prescript{}{B}{\mathbf{v}}_\infty}_{m}$ subject to Gaussian noise $\boldsymbol{\eta}_\text{LSTM}\sim \mathcal{N}(\mathbf{0}, \boldsymbol{\Sigma}_\text{LSTM})$. The associated measurement covariance $\boldsymbol{\Sigma}_\text{LSTM}$ is fixed and identified experimentally. Its output can be mapped to the filter state via the function $h_\text{airflow}$, defined as: 
\begin{equation}
   {\prescript{}{B}{\mathbf{v}}_\infty}_m =  {\mathbf{R}_{W}^{B}}^\top\text{exp}(\boldsymbol{\phi}^{\wedge})(\prescript{}{W}{\mathbf{w}} - \prescript{}{W}{\mathbf{v}} ) + \boldsymbol{\eta_\text{LSTM}}
   \label{eq:rel_airflow_in_body}
\end{equation}
where $\prescript{}{W}{\mathbf{w}}$ is defined in \cref{eq:wind}. The associated measurement noise vector is given by the vectors $\boldsymbol{\mathbf{\eta}}_\text{airflow} = \{ \boldsymbol{\eta}_\text{LSTM},  \boldsymbol{\eta}_m \}$. 

\paragraph{Unreliable odometry estimator measurement model}
The Unreliable Odometry Estimator provides position $\prescript{}{W}{\mathbf{p}}_m$, velocity $\prescript{}{W}{\mathbf{v}}_m$ and attitude ${\mathbf{R}_W^B}_m$ measurements subject to white noise, expressed with respect to the inertial reference frame $W$. Its output can be mapped to the state vector with a linear measurement update, provided that the measured attitude is transformed to the error-state representation via ${\boldsymbol{\phi}_m} = \log({\mathbf{R}_{W}^{B}}_\text{ref}{\mathbf{R}_W^B}_m^{-1})^\vee$. 

\subsection{Estimation scheme}
 The state of the filter is estimated using a discrete-time \ac{EKF}, where the prediction model is discretized using the forward Euler integration scheme. We note that the \ac{GP} used to represent the wind map can be incorporated in the filter by following the approach in \cite{ko2009gp}. The non-additive noise term given by the \ac{GP} (in \cref{eq:rel_airflow_in_body} due to \cref{eq:wind}) can be taken into account via the non-additive measurement noise variant of the EKF~\cite{simon2006optimal}.

\section{WIND MAP}
\label{sec:wind_mapping}
This section describes the approach used to represent and generate a probabilistic map of the velocity of the wind, used in our approach to perform dead reckoning in environments with stationary (position-dependent) wind.

\paragraph{Wind map model and assumptions}
The goal of the wind map $\hat{\mathcal{M}}(\cdot)$ is to represent a stationary, spatially continuous, 3D wind field $\mathcal{M}:\mathcal{E} \to \mathbb{R}^3$ in the environment $\mathcal{E} \subset \mathbb{R}^3$.
The wind map $\hat{\mathcal{M}}$ is generated from $K$ noisy wind velocity measurements $\hat{\mathbf{w}}_{1:K}$ and their corresponding positions $\hat{\mathbf{p}}_{1:K} \in \mathcal{E}$. Such measurements can be obtained from the proposed Airflow-Inertial Odometry estimator whenever position information is available, 
and by setting the wind map in \cref{eq:wind} to be empty, with a fixed covariance. 
We assume that the so obtained wind estimates $\hat{\mathbf{w}}_{1:K}$ (treated as input measurement by the map) are corrupted by white noise. While, in practice, turbulence or estimation errors (e.g. due to linearization) may cause correlation, the independence assumption can be better enforced by sub-sampling at a low frequency. Since the map is estimated via measurements collected when the Unreliable Odometry Estimator is available, we assume that the sequence of position estimates $\hat{\mathbf{p}}_{1:K} \in \mathcal{E}$ corresponding to $\hat{\mathbf{w}}_{1:K}$ is perfectly known. %

\paragraph{Wind map representation using Gaussian Processes}
Following \cite{lawrance2010simultaneous, popovic2020informative}, the presented assumptions allow us to represent the wind map $\hat{\mathcal{M}}(\cdot)$ using \acp{GP} \cite{rasmussen2003gaussian}. A \ac{GP} maps a position $\mathbf{p} \in \mathcal{E}$ to a mean wind velocity $\mathcal{M}(\mathbf{p})$ and an associated covariance $\boldsymbol{\Sigma}_\mathcal{M}(\mathbf{p})$ (assuming a Gaussian distribution), so that the underlying wind field is continuous and  differentiable.
In our context, \acp{GP} are used to describe distributions over wind fields given the training data $\mathcal{D} =\{(\hat{\mathbf{p}}_k, \hat{\mathbf{w}}_{k})\}_{k=1}^{K}$, which represent noisy observations of the underlying wind field $\mathcal{M}$. Specifically, a \ac{GP} is initialized with a zero-mean prior $p(\mathcal{M})$ on the wind field that we seek to estimate, and the training data is then used to induce a posterior $\hat{\mathcal{M}} \sim p(\mathcal{M}|\mathcal{D})$.  
To define a \ac{GP}, we need to select a covariance function (or kernel) $k(\mathbf{p}, \mathbf{p}')$, which acts as a prior on the generalization properties and the smoothness of the underlying wind field,  taking into account the spatial correlation between wind measurement taken at positions $\mathbf{p}$ and $\mathbf{p}'$. We choose a \ac{RBF} kernel, as it is commonly used to describe wind \cite{lawrance2010simultaneous, yang2017real}.
Since \acp{GP} operate on real valued functions, while we seek to predict an output $\mathbf{w} \in \mathbb{R}^3$, we employ three separate \acp{GP} $\hat{\mathcal{M}} = [\hat{\mathcal{M}}_x, \hat{\mathcal{M}}_y, \hat{\mathcal{M}}_z]^T$, one for each component of the wind field, expressed in the inertial reference frame $W$.

\paragraph{Sparse Gaussian Process for computational efficiency}
Given a training set of size $K$, a \ac{GP} can be trained with computational cost $O(K^3)$, and it can predict the value of the wind at an unseen location with cost $O(K)$. In order to reduce the computational complexity, allowing our approach to handle large maps and to perform computationally-efficient queries (since $K$ can be large), we utilize the Sparse \ac{GP} approach proposed in \cite{titsias2009variational}. A Sparse \ac{GP} identifies a subset $\mathcal{Z} \subset \mathcal{E}$ of size $M$ (with $M \ll K$) of the data (called inducing points) to be used for training and inference. The computational cost can thus be reduced to $O(KM^2)$ for training and $O(M)$ for inference of a new datapoint. 
The set of inducing points $\mathcal{Z}$, as well as the hyper-parameters of the kernel, are found by training the \ac{GP} offline.

\section{EVALUATION}
\label{sec:evaluation}
This section describes the strategy used to train the LSTM, and evaluates the proposed dead reckoning approach by simulating failures of the Unreliable Odometry Estimator in two scenarios, with and without wind in the environment.

\subsection{Datasets collection}
\label{sec:evaluation:datsets}
In order to evaluate the proposed approach and generate training data for the LSTM, we collect datasets by flying indoor an hexarotor, shown in \cref{fig:intro_figure}, equipped with four airflow sensors. Inertial data is collected at 200Hz from the onboard IMU (MPU-9250 \cite{MPU9250T56:online}). 
The ground-truth pose of the robot is provided by a motion capture system, while ground-truth velocity information is obtained by an estimator running onboard, which fuses the pose information with the inertial data from an IMU.
The airflow-data is obtained at 50Hz, synchronously, from the four airflow sensors. 
We use the MIT/ACL open-source snap-stack \cite{acl_snap_stack} for controlling the MAV. 
Wind is generated via an array of five leaf-blowers and a large fan, shown in \cref{fig:exprimental_setup}. They are pointed towards the same direction, and they are set to produce approximately 2m/s wind speed at a distance of 3 meters.   

\subsection{LSTM architecture, training, and ablation study}
As in our previous work \cite{tagliabue2020touch}, the architecture of the network is based on a 2 hidden layers LSTM, with 16 nodes per layer, and a sequence length of five steps. The network is trained from 8 minutes of data collected by flying indoor without wind, assuming that the measured velocity of the robot $\prescript{}{B}{\mathbf{v}}$ corresponds to the relative airflow $-\prescript{}{B}{\mathbf{v}}_\infty$ detected by the sensors. %
Training is performed using the ADAM optimizer, while minimizing the MSE loss. 

We performed an ablation study to evaluate the relevance of adding throttle and accelerometer data as inputs to the network. The results in \cref{tab:ablation_study} confirm that both inputs contribute to improving the accuracy of the network. Using the two inputs, however, yields marginal improvements compared to using only one of the two. This may be caused by the fact that accelerometer and throttle data contain redundant information.
\begin{table}[t]
    \vspace*{0.1in}
    \caption{Relative changes in Mean-Square error on the test set when varying the inputs signals used for training of the LSTM network. The results highlight the importance of adding either accelerometer or throttle data as input of the network.}
    \label{tab:ablation_study}
    \centering
    \begin{tabular}{l|c}
        Inputs & Rel. change in MSE loss \\
        \hline 
        \hline 
        Airflow, Gyro., Throttle, Acc. (baseline) & -  \\ %
        Airflow, Gyro., Throttle & $2.28 \%$ \\ %
        Airflow, Gyro., Acc. & $3.90 \%$ \\ %
        Airflow, Gyro. & $13.00 \%$ \\ %
    \end{tabular}
    \vspace{-0.1in}
\end{table}

\subsection{Wind map generation}
\label{sec:evaluation:wind_map}
\begin{figure}[t]
    \centering
    \vspace*{0.05in}
    \includegraphics[trim = 0 0 40 0,clip,width=1.0\columnwidth]{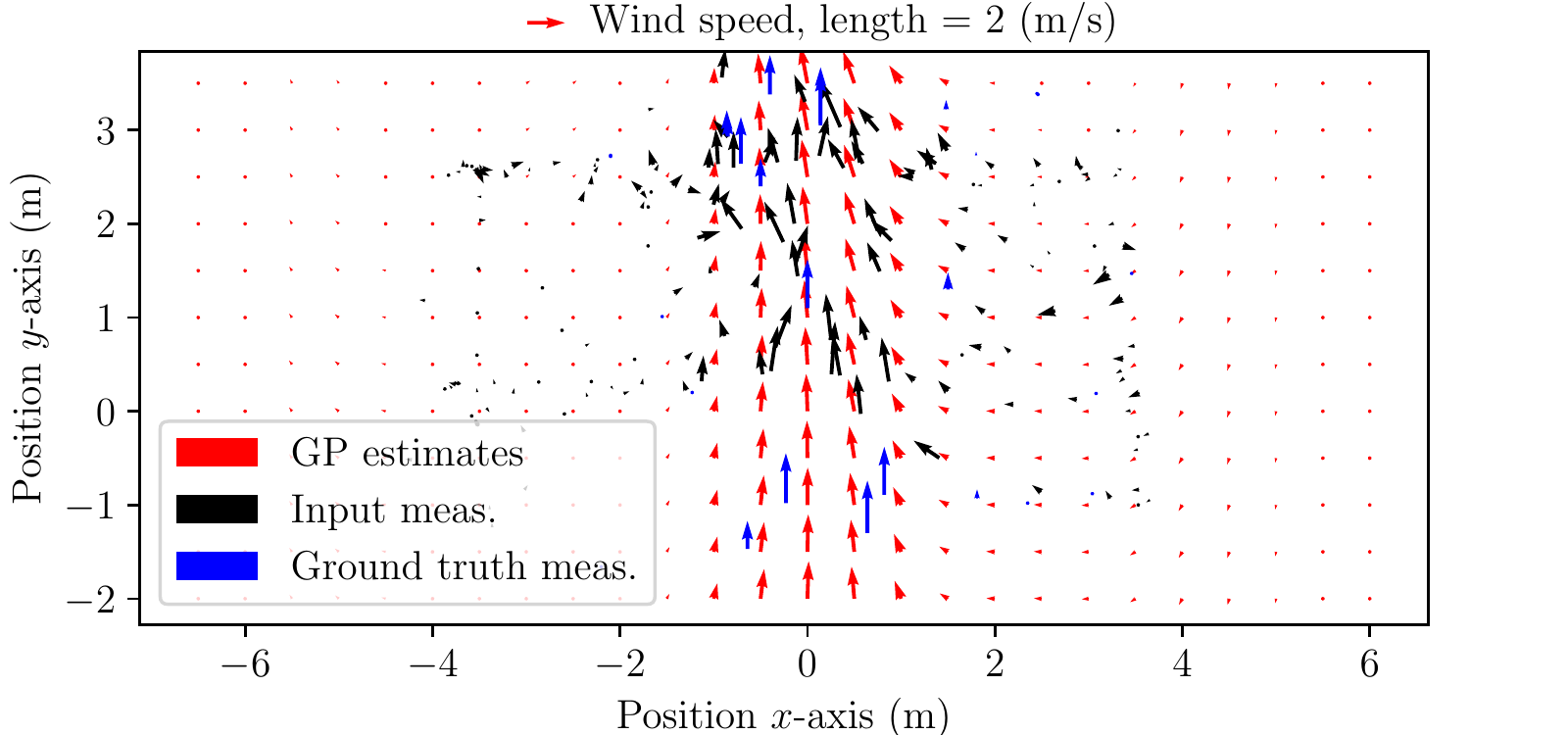}
    \caption{Map of the wind generated using an array of five leaf-blowers and a large fan. The map (red arrows) is generated via a Sparse Gaussian Process (GP) from the wind estimated onboard (black arrows). Ground truth wind speed measurements (blue arrows) were collected with an hot-wire anemometer. Red arrow at top of the plot denotes the scale corresponding to a 2m/s wind speed.}
    \label{fig:wind_map_xy}
    \centering
    \includegraphics[trim = 240 150 40 0,clip,width=0.75\columnwidth]{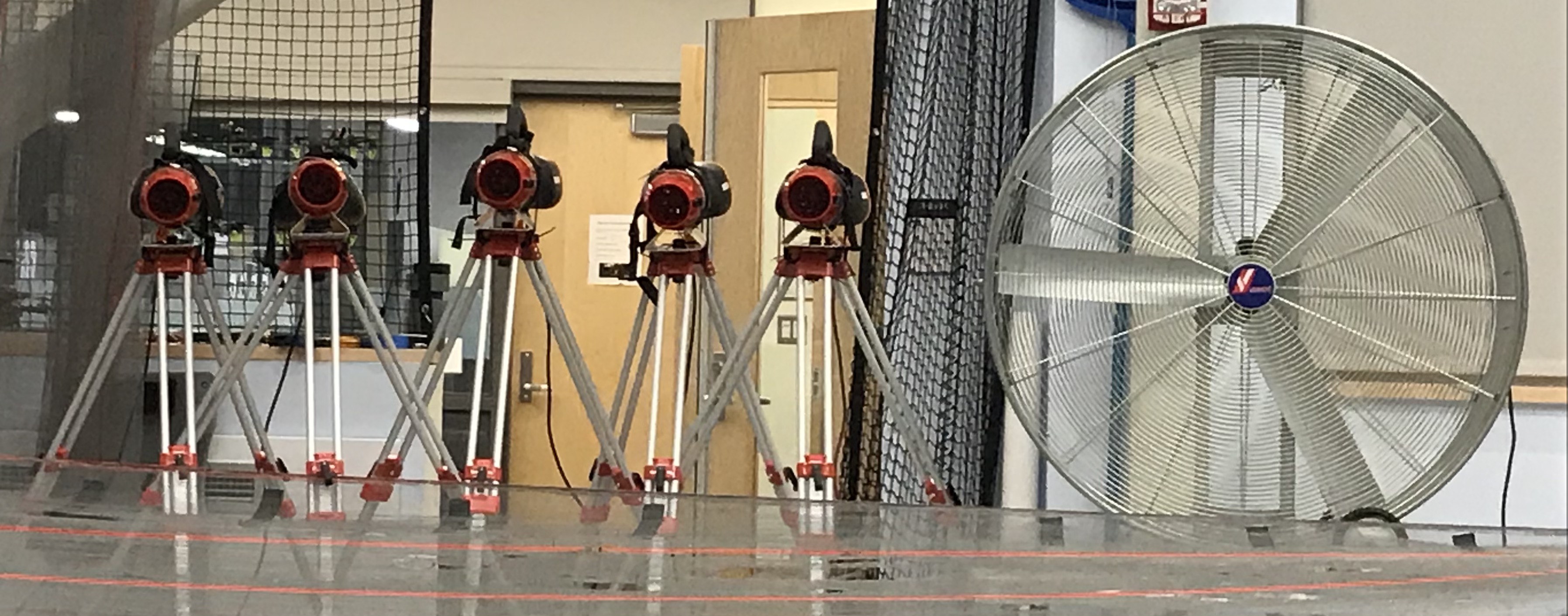}
    \caption{Experimental setup used to generate wind. }
    \label{fig:exprimental_setup}
    \vskip-3ex
\end{figure}

We generate a 3D map of the wind speed in the flight space under the setup shown in \cref{fig:exprimental_setup}. We fly in front of the fan and the array of leaf-blowers multiple times, in different directions and orientations, for a total of approximately two minutes. The wind $\prescript{}{W}{\hat{\mathbf{w}}}$ is estimated via the proposed EKF by setting the wind map model to be empty, with a fixed positive  covariance $\boldsymbol{\Sigma}_{w}$, and using position and attitude measurements from the motion capture system. The map is generated offline from the collected wind-estimates $\hat{\mathbf{w}}_{1:K}$, sub-sampled to 1Hz in order to reduce effects of correlation due to the turbulence in the wind. 
The Sparse \acp{GP} framework described in Section \ref{sec:wind_mapping} is trained with $M = 20$ inducing points using the GPy library \cite{gpy2014}. The resulting wind map captures well the region with high wind produced by our setup, as can be seen by the projection on the $x$ and $y$ axis of the flight space (at an altitude of $z = 2$m), shown in \cref{fig:wind_map_xy}. The black arrows represent the sub-sampled wind-estimates $\prescript{}{W}{\hat{\mathbf{w}}}$ treated as input measurements for the map, while the red arrows represent the wind speed estimates obtained from the \acp{GP}.  
In order to validate the accuracy of the wind map, we separately collect measurements of the wind speed at about 50 different locations using an hot-wire anemometer, represented by the blue arrows in \cref{fig:wind_map_xy}, obtaining a RMSE of $0.67$m/s. The error is relatively high in part because multiple ground-truth measurements have been collected in proximity of the leaf blowers, where the robot did not directly fly and so did not collect any measurement.

\subsection{Evaluation metrics and details for odometry estimation}
We evaluate the performance of the dead reckoning approach by computing the following errors (following \cite{liu2020tlio}):
\begin{enumerate}[a)]
    \item \textbf{RMSE}: Root Mean Square error,
    here called \textbf{RMSE} for  position, and \textbf{RMSE-yaw} for  yaw.
    \item \textbf{DR}: total drift of trajectory and ground truth, obtained as $\|\mathbf{p}_N- \hat{\mathbf{p}}_N \|/(\text{total traj. length})$, where $N$ is the last time-step.
    \item \textbf{RTE-2s}: Compares Relative Translation Error of the trajectory to ground truth over a 2s window, by compensating for the effects of the drifts on yaw, via $\sqrt{\frac{1}{N}\sum_{i}\| \mathbf{p}_{i+2s} - \mathbf{p}_{i} - \mathbf{R}_{\text{yaw}}\mathbf{\hat{R}}_{\text{yaw}}^T(\hat{\mathbf{p}}_{i+2s} - \hat{\mathbf{p}}_{i})\|^2}$
\end{enumerate}
Note that the estimation performance of roll and pitch angles is better or comparable to IMU-only dead reckoning.
The algorithms are evaluated on an Intel i7 laptop equipped with an Nvidia RTX 2070 GPU. The AIO Estimator (implemented in C++) runs at 200Hz using approximately 10\% of one core of the CPU, while the wind map (Python) runs at 25Hz using one core of the CPU. The Relative Airflow Estimator is set to run at 25Hz on the GPU.

\subsection{Evaluation with constant, zero wind}
\begin{figure}[t]
    \centering
    \vspace*{0.05in}
    \includegraphics[trim=0 0 0 0,clip,width=0.95\columnwidth]{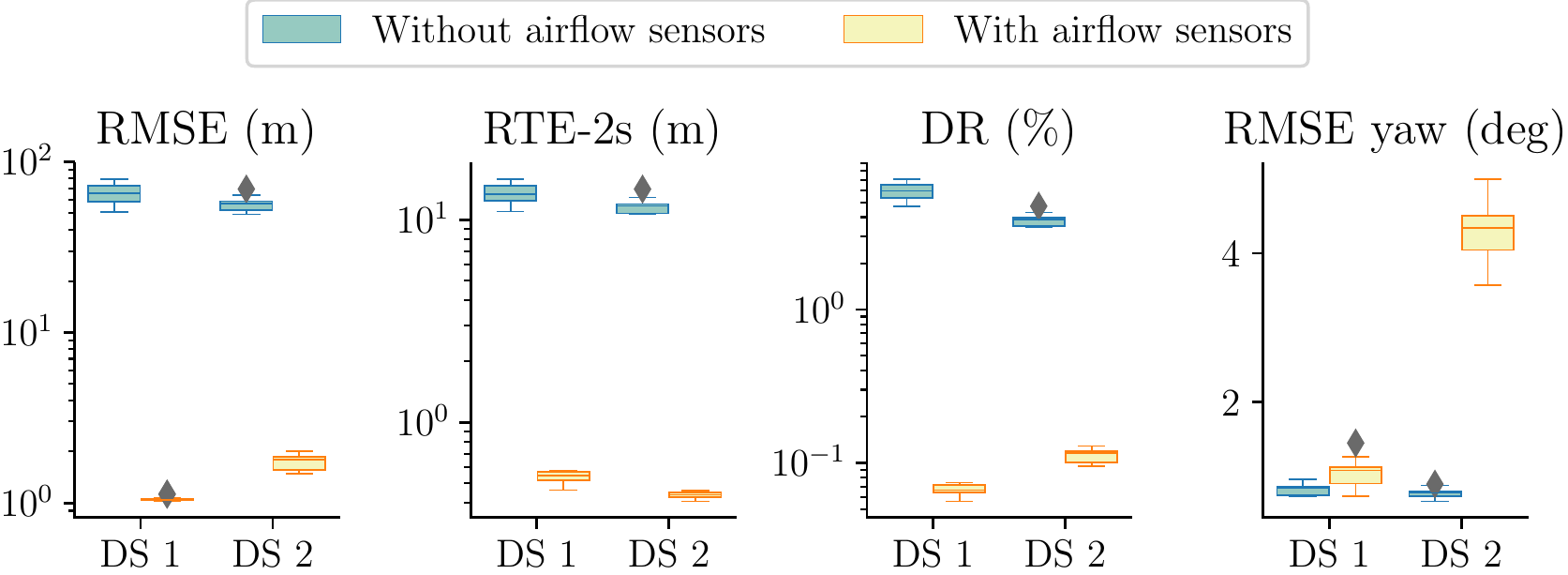}
    \caption{Comparison of different error metrics for the position and rotation of IMU-only dead reckoning and Airflow-Inertial dead reckoning (our proposed strategy) over a 30s long trajectory. Please note the logarithmic scale on the $y$ axes. Our approach achieves up to one order of magnitude improvements in the Root Mean Square error (RMSE), relative translation error over a 2s window (RTE-2s), and the total drift (DR). Yaw estimation performance (RMSE-yaw) is comparable with IMU-only dead reckoning.}
    \label{fig:boxplot_no_wind}
\end{figure}

\begin{figure}
\captionsetup[sub]{font=footnotesize}
\centering
\begin{subfigure}{\columnwidth}
  \centering
  \includegraphics[trim=0 0.5 0 0,clip,width=\columnwidth]{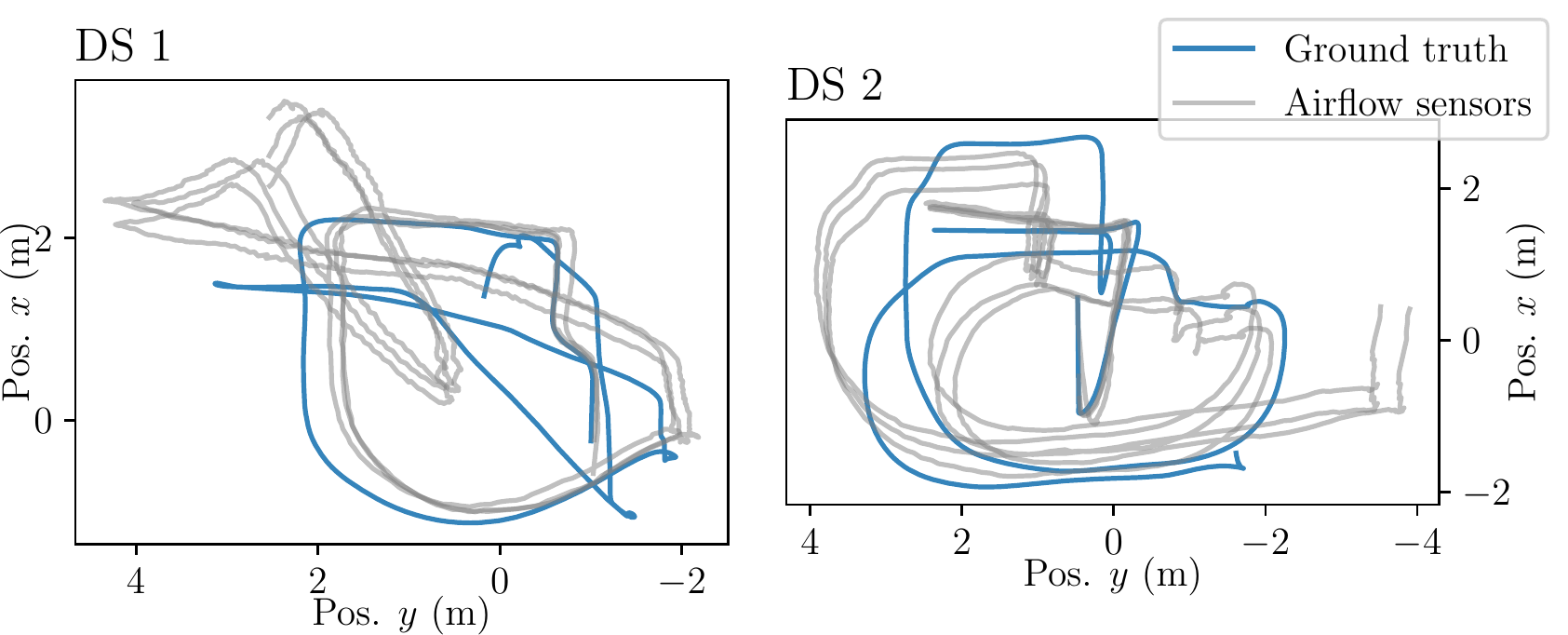}
  \caption{DS 1 and DS 2, without wind. The total time of both trajectories is 30s. }
  \label{fig:traj_no_wind}
\end{subfigure}%
\hspace{0.1cm}
\begin{subfigure}{\columnwidth}
  \centering
  \includegraphics[trim=0 0.5 0 0,clip,width=0.92\columnwidth]{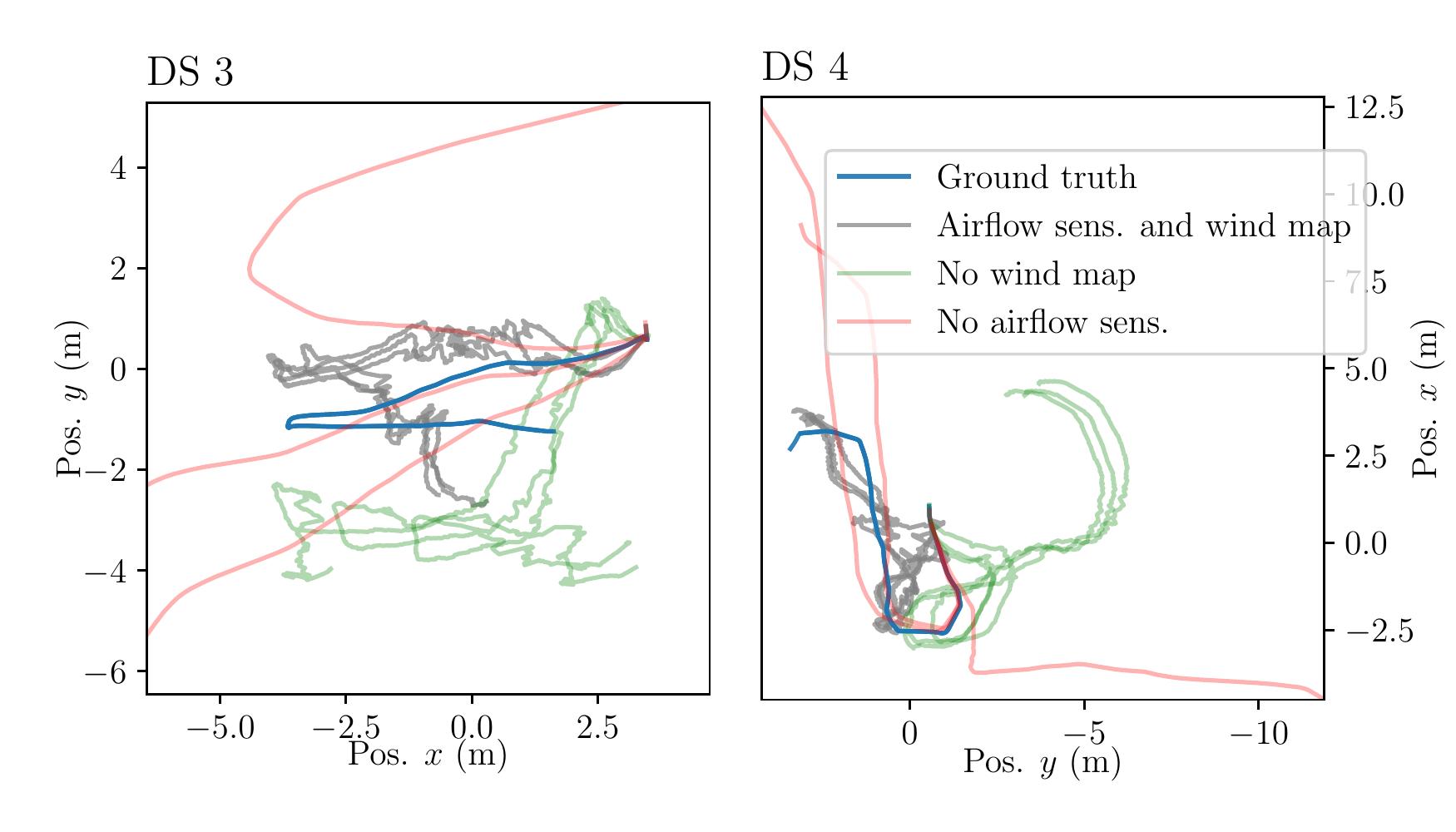}
  \caption{DS 3 and DS 4, with wind. The trajectories are 25s and 20s long, respectively. }
  \label{fig:traj_with_wind}
\end{subfigure}
    \caption{Ground truth and estimated trajectories on the x and y axes of the different datasets. DS~1 and 2 present aggressive maneuvers with speed up to 2.5m/s and motions around yaw. DS~3 and 4 represent the UAV transitioning back and forth into the high wind area produced by the leaf-blowers/fan. The wind blows towards the +y direction. The difference in estimated trajectories is caused by different convergence values of the biases and the wind when the failure is triggered.}
    \label{fig:trajectories}
    \vskip-2ex
\end{figure}

This part evaluates the accuracy of the proposed airflow-aided dead reckoning strategy in the case of constant, zero wind. We consider two distinct datasets (DS 1 and DS 2), of similar complexity. On each dataset we simulate a failure of the Unreliable Odometry Estimator (based on a motion capture system) after 20s of flight, and we compute the proposed evaluation metrics after 30s of flight. We repeat the experiment for ten times, randomly triggering the failure within a two seconds window after the 20s of initial flight.
We compare our strategy with IMU-only dead reckoning, obtained by fusing only IMU-data in the proposed EKF. 
The results are presented in \cref{fig:boxplot_no_wind}, and show that our approach consistently provides benefits in terms of reduced position drift and closeness of the estimated trajectory to ground truth, as captured by the RMSE and RTE-2s, with a reduction of up to one order of magnitude when compared to the IMU-only strategy. Drift in yaw remains comparable to the IMU-only strategy, since the airflow sensors do not provide angular velocity information in absence of wind. \cref{fig:traj_no_wind} shows ground truth and three estimated trajectories using the proposed methods, and demonstrates that our approach well captures the motion of the robot, with very limited drift. \cref{fig:intro_figure} shows trajectories from dataset 2, and it highlights (a) the large drift of the IMU-only strategy, and (b) the small altitude drift of our strategy.

\begin{figure}[t]
    \centering
    \vspace*{0.05in}
    \includegraphics[width=0.95\columnwidth]{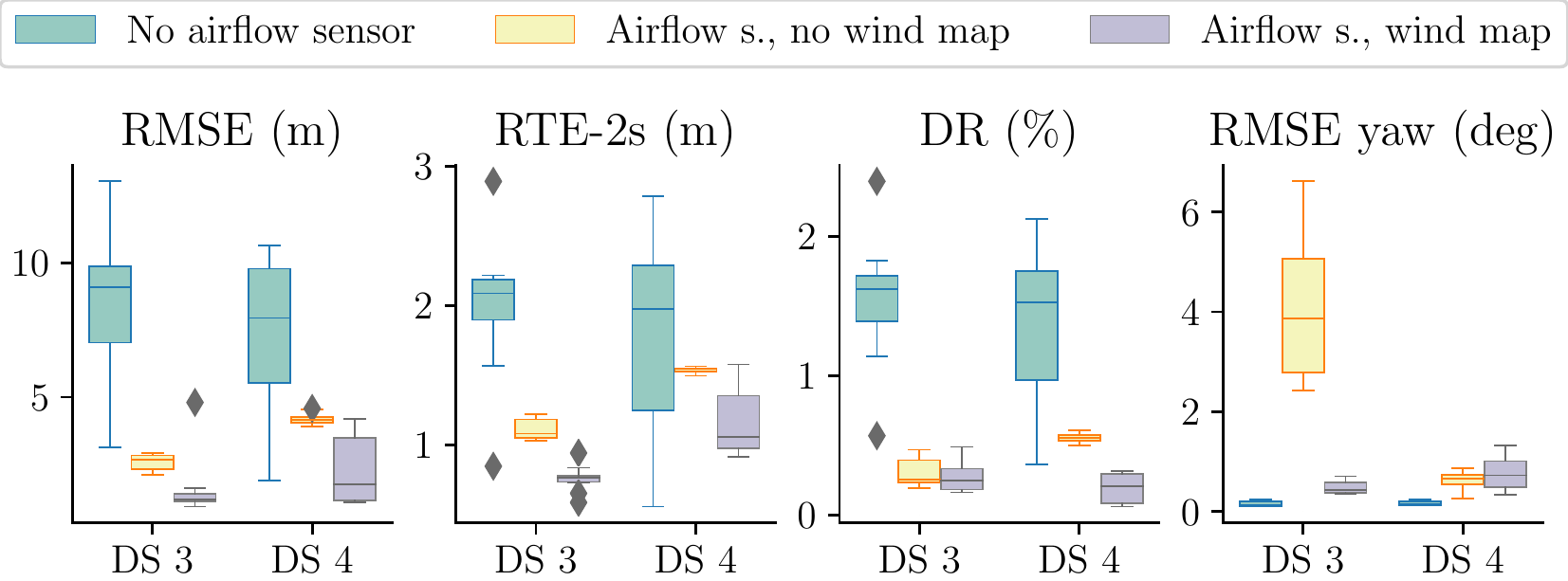}
    \caption{Evaluation of the performance of the proposed approach in the scenario of position-dependent wind, generated from an array of leaf-blowers and a large fan. The results show that the 25s- (for DS 3) and 20s-long trajectories (for DS 4) estimated while crossing a windy region (IMU + airflow sensors + wind map) more closely follows the ground truth then the compared strategies (IMU only and IMU + airflow sensors, but without a map of the wind). This is confirmed by the Root Mean Square error (RMSE), relative translation error over a 2s window (RTE-2s), and the total drift (DR), computed over three tests. Yaw estimation performance (RMSE-yaw) is comparable with respect to IMU-only dead reckoning.}
    \label{fig:boxplot_with_wind}
    \vskip-2ex
\end{figure}

\subsection{Evaluation with non-constant, stationary wind}
This part evaluates the accuracy of the proposed strategy by relaxing the assumption of constant wind, allowing the wind to be spatially varying.
Wind is generated experimentally according to the procedure detailed in \cref{sec:evaluation:datsets} and the setup shown in \cref{fig:exprimental_setup}, and its effects are compensated in the filter via the wind map generated according to \cref{sec:evaluation:wind_map}. We note that the wind map is generated via measurements collected in a separate experiment, following distinct trajectories from the one used for evaluation of the approach.
A failure of the Unreliable Odometry Estimator is triggered after 20s of flight, while the UAV is in an area with zero wind, and we evaluate the estimated trajectories as the robot transitions back and forth the region with wind, as shown in \cref{fig:traj_with_wind}, corresponding to 25s and 20s of flight for dataset 3 and 4, respectively.  
The evaluation was repeated ten times, triggering the odometry failure within a two-second window.
The results in \cref{fig:traj_with_wind} and \cref{fig:boxplot_with_wind} include a comparison with IMU-only dead reckoning and with the proposed strategy without a wind map.
\cref{fig:traj_with_wind} shows that the estimated trajectory more closely follows the ground truth, without being subject to the large drift present in the compared strategies. The estimated trajectories, however, are less smooth than in the other approaches, due the correlated noise caused by the turbulent wind and the position updates enabled by the wind map. In addition, as the robot drifts in the map, the ability to compensate for the wind effects decreases, which limits the approaches effectiveness to short range trajectories. The results in \cref{fig:boxplot_with_wind} still demonstrate that the wind map contributes to reducing the drift (DR) when compared to IMU-only dead reckoning, and the estimated trajectories achieve higher spatial closeness (RMSE and the RTE-$2s$) to ground truth. The performance in yaw estimation is comparable to IMU-only dead reckoning.

\section{CONCLUSION AND FUTURE WORK}
\label{sec:conclusion}
This work presented a strategy to perform dead reckoning using inertial data and 3D airflow measurements. %
From data collected experimentally, we have shown that airflow-aided dead reckoning can effectively reduce drift (up to one order of magnitude over a 30 seconds trajectory) with respect to an IMU-only solution in non-windy environments. We have additionally shown that utilizing a map of the wind helps to reduce drift and position errors, even in turbulent wind environments.%
The results provide many opportunities for future work. Investigation into approaches to generate or update the wind map online would increase resilience in non-stationary wind fields, and employing more sophisticated filtering techniques may enable larger-scale localization. In addition, incorporating a drag and dynamics model of the robot may further constrain the estimation problem. %

\section*{ACKNOWLEDGMENT}
This work was funded by the Air Force Office of Scientific Research MURI FA9550-19-1-0386. 
The authors thank Austen J. Roberts for the ablation study on the LSTM.

\balance
\bibliographystyle{IEEEtran}
\bibliography{bibiliography.bib}

\end{document}